\documentclass[conference]{IEEEtran} 
\usepackage{algorithm}
\usepackage{pdfpages}
\usepackage{epstopdf}
\AppendGraphicsExtensions{.tif}
\usepackage[inkscapeformat=png]{svg}
\usepackage{tabularx}
\usepackage{amssymb}
\usepackage{graphicx}
\usepackage{subcaption}
\usepackage{url}
\usepackage{cite}
\usepackage{amsmath,amssymb,amsfonts}
\usepackage{algorithmic}
\usepackage{graphicx}
\usepackage{textcomp}
\usepackage{xcolor}
\def\BibTeX{{\rm B\kern-.05em{\sc i\kern-.025em b}\kern-.08em
    T\kern-.1667em\lower.7ex\hbox{E}\kern-.125emX}}

\usepackage{comment}

\ifCLASSOPTIONcompsoc
  \usepackage[nocompress]{cite}
\else
  \usepackage{cite}
\fi
\ifCLASSINFOpdf
\else
\fi
\makeatletter
\newcommand{\linebreakand}{%
  \end{@IEEEauthorhalign}
  \hfill\mbox{}\par
  \mbox{}\hfill\begin{@IEEEauthorhalign}
}
\makeatother

\begin{document}
%
\title{Supersampling of Data from Structured-light Scanner with Deep Learning}

\author{\IEEEauthorblockN{1\textsuperscript{st} Martin Melicherčík}
\IEEEauthorblockA{\textit{Department of Applied Informatics} \\
\textit{Comenius University}\\
Bratislava, Slovakia \\
martin.melichercik@gmail.com}
\and

\IEEEauthorblockN{2\textsuperscript{nd} Lukáš Gajdošech}
\IEEEauthorblockA{\textit{Department of Applied Informatics} \\
\textit{Comenius University}\\
Bratislava, Slovakia \\
lukas.gajdosech@fmph.uniba.sk\\
0000-0002-8646-2147}
\and

\IEEEauthorblockN{3\textsuperscript{rd} Viktor Kocur}
\IEEEauthorblockA{\textit{Department of Applied Informatics} \\
\textit{Comenius University}\\
Bratislava, Slovakia \\
viktor.kocur@fmph.uniba.sk\\
0000-0001-8752-2685}
\and
\linebreakand
\IEEEauthorblockN{4\textsuperscript{th} Martin Madaras}
\IEEEauthorblockA{\textit{Department of Applied Informatics} \\
\textit{Comenius University}\\
Bratislava, Slovakia \\
martin.madaras@fmph.uniba.sk\\
0000-0003-3917-4510}
\IEEEauthorblockA{\textit{Skeletex Research} \\
Bratislava, Slovakia \\
madaras@skeletex.xyz}
\vspace{-9mm}
}

\IEEEoverridecommandlockouts
\IEEEpubid{\makebox[\columnwidth]{979-8-3503-4353-3/23/\$31.00~\copyright2023 IEEE\hfill} \hspace{\columnsep}\makebox[\columnwidth]{ }}

\maketitle
\IEEEpubidadjcol
\begin{abstract}

This paper focuses on increasing the resolution of depth maps obtained from 3D cameras using structured light technology. Two deep learning models FDSR and DKN are modified to work with high-resolution data, and data pre-processing techniques are implemented for stable training. The models are trained on our custom dataset of 1200 3D scans. The resulting high-resolution depth maps are evaluated using qualitative and quantitative metrics. The approach for depth map upsampling offers benefits such as reducing the processing time of a pipeline by first downsampling a high-resolution depth map, performing various processing steps at the lower resolution and upsampling the resulting depth map or increasing the resolution of a point cloud captured in lower resolution by a cheaper device. The experiments demonstrate that the FDSR model excels in terms of faster processing time, making it a suitable choice for applications where speed is crucial. On the other hand, the DKN model provides results with higher precision, making it more suitable for applications that prioritize accuracy.
\end{abstract}


%
\IEEEpeerreviewmaketitle

\section{Introduction}
In recent years, there has been a growing interest in devices that provide 3D information about captured scenes, alongside traditional 2D image-capturing cameras. Along with depth maps, these devices also provide texture images of the scenes. Working with 3D data has numerous applications, including validating components, analyzing archaeological findings, and enabling robotic tasks such as mapping and orientation in space. 
However, processing depth maps can be computationally demanding, especially when real-time processing is required. The higher the resolution of the depth map, the longer it takes to process. To address this challenge, this paper proposes a method that involves down-sampling the depth map to a lower resolution, processing it at that resolution, and then up-sampling it back to a high resolution. This approach aims to reduce the computational complexity while still maintaining the quality of the resulting depth map.

The idea of up-sampling processed depth maps originates from the video game industry, where game producers aim to provide players with lower GPU computing power with a higher resolution and frame rate experience. To achieve this, various technologies based on deep learning techniques have been developed, with DLSS (Deep Learning Super Sampling) \cite{dlss} by NVIDIA is one of the most well-known examples. Inspired by this approach, our task will be to take the input depth map, down-sample it, apply the filters and processing we need, and then up-sample the depth map back to high resolution preserving its sharp edges and surface details. Numerous studies have explored this task, with deep learning models emerging as a popular solution. 
In this paper, we examine two deep learning models, DKN \cite{dkn} and FDSR \cite{fdsr}, and modify them to work with high-resolution and accurate data obtained from the motion MotionCam-3D device \footnote{\url{https://www.photoneo.com/motioncam-3d/}}. These models learn to up-sample depth maps while preserving sharp scene details. Additionally, the paper addresses the issue of data pre-processing to ensure stable training of the modified models. It also evaluates the resulting depth maps using qualitative and quantitative metrics, focusing on the quality of the 3D image produced. Furthermore, the time efficiency and usability of the proposed solution are also evaluated.

The contributions of this paper lie in the development of a method to increase the resolution of depth maps using deep learning models and the exploration of its applications in various 3D tasks. By improving the resolution of depth maps, we aim to enhance the overall quality of 3D data processing and enable more efficient and accurate analysis in application fields. Neural network models and training data are available on GitHub.\footnote{\url{https://github.com/Meli-0xFF/depthmap_sr}}

\subsection{Problem Definition}

Let us define the problem we are solving formally. The input data will be paired $D_{HR}, I_{HR}$ that stands for high-resolution depth map and intensity texture of the scanned scene. Let $W, H$ be the width and height of the input images in pixels. 
We will consider down-sampling as a procedure that gets depth map $D_{HR}$ with resolution $H \times W$ pixels as input and gives depth map $D_{LR}$ with resolution $H / s \times W / s$ where constant $s \in \mathbb{N}$ will be called a down-sample factor. The down-sample factor $s$ determines the size of the $s \times s$ pixels square in $D_{HR}$ that will be compressed to one pixel in $D_{LR}$. The low-resolution depth map $D_{LR}$ will be input for our main task: depth map up-sampling. Additional data that can help us transform a low-resolution depth map into a high-resolution one is the high-resolution intensity texture from the scanner. Output of up-sampling procedure should be depth map $O$ that has resolution $H \times W$ (same as input $D_{HR}$).\\

\section{Available Training Data}
There are few available datasets that contain pairs of the depth map and RGB or intensity texture. They differ in depth map creation methods. Examples of such datasets are NYUv2 \cite{nyuv2}, Middlebury \cite{middlebury}, and RGB-D-D \cite{fdsr}. The NYUv2 dataset \cite{nyuv2} was created using Microsoft Kinect \cite{kinect}, and its samples are scans from video sequences of interior spaces like kitchens, offices, living rooms, and others. The Middlebury \cite{middlebury} dataset provides several sets of data. Some are generated by stereo vision, others with a structured light scanner. Dataset sample scenes contain various arranged objects with complex surfaces. The last mentioned RGB-D-D \cite{fdsr} dataset was created using two devices: a ToF camera and an RGB camera. Textures and depth maps are aligned and have the exact resolution obtained by cropping the window from the texture. 

All three mentioned datasets have one property in common: they are densely defined. Depth maps in these datasets contain only small non-defined holes or areas that are simple to fill in from their surrounding defined pixels. Objects in scenes of datasets created by a ToF camera or Kinect do not have very sharp edges, and their samples are generally quite noisy. 

\subsection{Photoneo MotionCam-3D}
\begin{figure}
    \begin{subfigure}{.49\columnwidth}
      \includegraphics[width=0.95\columnwidth]{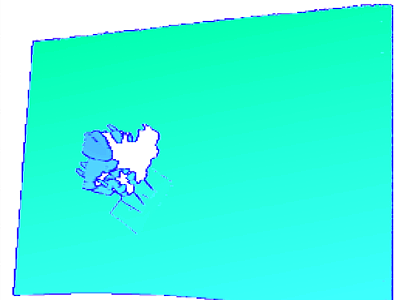}
    \end{subfigure}%
    \begin{subfigure}{.49\columnwidth}
      \includegraphics[width=0.95\columnwidth]{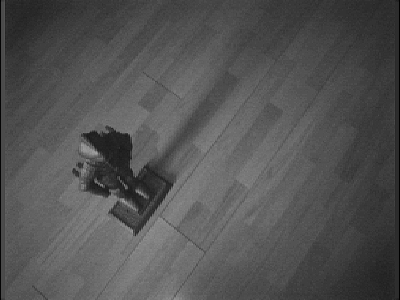}
    \end{subfigure}
    \caption[MotionCam-3D samples]{Samples from Photoneo MotionCam-3D. The depth map (left) is visualized in RGB and mapped linearly to the color scale from green to blue. White pixels are undefined. The intensity map is visualized on the right.}
    \label{fig:scanner_data}
\end{figure}

To create our dataset, we utilized the Photoneo MotionCam-3D, a 3D camera based on structured-light technology. Considering the use case of MotionCam-3D, which is commonly used for 3D object fusion tasks, we designed our dataset scenes to align with this specific application.
The scenes were carefully composed to ensure that the objects had relatively complex surfaces, allowing for a challenging and realistic representation of the fusion task. 
The structured-light computation results in undefined pixels in the depth map. The shadows cast by objects can create regions where the pattern is not visible or distorted, leading to missing or unreliable depth information. In order to deal with the missing depth information the depth maps have to be carefully preprocessed. Intensity texture from the scanner camera can provide helpful information about undefined regions. For illustration, we present samples from scanner data in Fig. \ref{fig:scanner_data}. We can observe that the depth map contains a lot of undefined (white) pixels. 

We selected five objects for our dataset, each with unique characteristics: a warrior figure, a 3D-printed bear figure, a David statue model, an HDD disc, a rotor, and a cogwheel. By including objects with diverse shapes and surface textures, we aim to provide a varied and comprehensive dataset for our experiments. In total, our dataset comprises 1200 samples, each representing a scene with a single object on a flat surface. The scenes were captured using the Photoneo MotionCam-3D, ensuring that the depth maps and intensity textures were aligned accurately. An example of a scene from our dataset is in Fig. \ref{fig:scanner_data}, illustrating the object placed on the flat surface.

\section{Related Work}



In this section we describe different upsampling methods which we have modified and tested within our pipeline. Based on the results of a recent survey of various depth map super-resolution models \cite{dct} we have selected two models: DKN \cite{dkn} and FDSR \cite{fdsr}.


\subsection{DKN model}
The DKN architecture \cite{dkn} follows an image-filtering approach, where the input depth map is up-sampled using a simple method, and a custom filter is learned to enhance its details. This approach enables DKN to generate more accurate depth maps from the expanded, inexact depth map.

The DKN (Deformable Kernel Network) model, as described in its publication \cite{dkn}, is an approach for depth map super-resolution that is based on the concept of joint image filtering. The main idea behind this approach is to expand the image using a simple up-sampling method and then apply a filter to enhance the precision of the depth map. Applying the filter involves sliding a kernel window along the image. The kernel window determines the set of neighboring pixels for each pixel. By considering the values of neighboring pixels, a more accurate value can be computed for each pixel. The contribution of each neighboring pixel is determined by weight, such as a weighted average of neighboring pixels. It is important to note that the weights of the kernel pixels should sum to zero. Each pixel's kernel window weight values can be determined by guidance image (in our case, intensity texture) or just from the depth image structure. 
The DKN model uses the CNN model for picking neighbor pixels of each pixel and their weights. 
The model comprises two main branches: texture and depth. The network can be divided into three consecutive parts: feature extraction, weight \& offset regression, and weighted average. 

DKN model was trained on NYUv2 dataset \cite{nyuv2}, which was created using a Kinect device \cite{kinect}. The depth maps from this dataset are in lower resolution and do not have as sharp edges as our Photoneo data. The model works with a pre-expanded depth image. The low-resolution depth map is firstly expanded by bi-cubic interpolation and then put into the model for further processing. 
For training, the DKN model uses $L_1$ loss function that provides evenly distributed attention to all pixels of depth maps when computing sample error.

\subsection{FDSR model}

In contrast to DKN, the FDSR model \cite{fdsr} adopts a continuous residual learning strategy that leverages the structure from intensity texture to enhance the depth image. The architecture includes both a texture branch and a depth branch, with the texture branch merging into the depth branch at specific network layers. This merging allows FDSR to effectively utilize the information from the texture image to improve the quality of the depth map.


The FDSR model is based on continuous learning considering high and low-frequency components of input images. High-frequency components of the image, such as edges, are areas that are more important in the image structure manner. 
Low-frequency components such as planes are less important. 
The FDSR model learns to extract high-frequency components of a scanned scene from high-resolution texture and uses this information while learning the depth map up-sampling procedure. FDSR model architecture has two main branches: \textit{High-Frequency} branch and \textit{Multi-scale Reconstruction} branch. The high-resolution texture is input for the former, and the low-resolution depth map for the latter branch. 

In summary, the design of a CNN model for depth map super-resolution involves answering high-level questions about the placement and method of the up-sampling layer. The choice of up-sampling method depends on the data and task-specific requirements. The best-performing models FDSR and DKN, need to be adapted to the Photoneo MotionCam-3D data and the challenges it presents. The up-sampling layer is a key component in all models and can be ML-based or algorithmic, each with its own advantages and considerations.

Other recent trends include attention mechanisms emphasizing important areas such as edges \cite{attention}. Different novel solution employs a neural network to generate alternative views, super-sampling the depth map by fusing these perspectives \cite{channel}. Lastly, multiple aligned images taken in varying lightning conditions can be leveraged as opposed to a single texture \cite{light}.


\section{Data Preparation}
For the training of a depth map super-resolution CNN model, we need a dataset with samples containing HR intensity texture, LR depth map, and HR depth map as the ground truth. 
Samples can also contain additional data that can help us during training. Here we describe the preparation of the dataset and the structure of its samples. We will start with the generation of low-resolution depth maps that must be included in each sample.

\subsection{Depth Map Down-sampling}
In the formal definition, 
we have defined down-sampling determined by the scale factor $s \in \mathbb{N}$ as collapsing $s \times s$ square blocks to one pixel. Various deterministic algorithms can do depth map down-sampling. Authors in the majority of publications use methods based on bi-cubic interpolation or the nearest neighbor approach. 
Based on \cite{deconv}, applying the nearest neighbor on data from a structured-light scanner is preferred, while bi-cubic interpolation is better suited for data from ToF devices. Inspired by this study, we used a method based on the nearest-neighbor approach \cite{downsample}, which determines the value of the down-sampled pixel based on whether the block contains notable changes in the gradient. 

\begin{figure}
    \begin{subfigure}{.49\columnwidth}
      \centering
      \includegraphics[width=0.8\columnwidth]{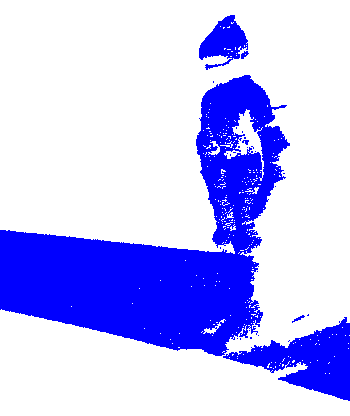}
    \end{subfigure}
    \begin{subfigure}{.49\columnwidth}
      \centering
      \includegraphics[width=0.8\columnwidth]{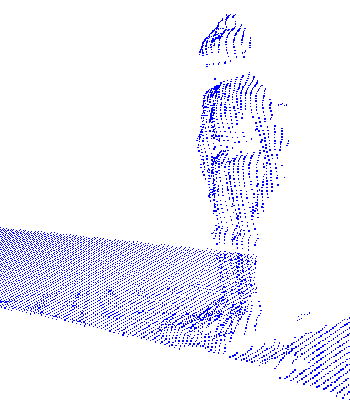}
    \end{subfigure}
    \centering
    \caption[Depth map down-sampling]{Point clouds from input HR depth map (left) and down-sampled LR depth map (right) with scaling factor $s = 4$.}
    \label{fig:downsample}
\end{figure}

Fig. \ref{fig:downsample} demonstrates the results of the down-sampling method. We computed 3D point clouds from the input high-resolution and output low-resolution depth map with scaling factor $s = 4$. We can observe that the depth map lost its density but not the sharpness of the object edges. 

\subsection{Depth Map Filling}
Both models that we want to use assume depth maps without holes, all pixels should be defined and contain meaningful, valid depth values. However, depth maps from the MotionCam-3D contain a large number of undefined pixels. The device often discard distant observations because of potential inexact values. We need to somehow treat these areas.
A simple approach based on ignoring the undefined regions in the loss computation leads to unstable training.
Therefore, we propose a filling procedure that gets a depth map with undefined regions as input and provides a filled depth map as the output. 
To preserve information about every sample's defined and undefined pixels, we introduce  a definition map. It is a 2D binary matrix with the dimensions of the HR depth map, which elements contain value 0 if the pixel at the same position in the HR depth map is not defined and value 1 if the corresponding depth map pixel is defined. As the depth map itself will be modified, the definition map provides stable information regarding the originally defined regions.

Now we will introduce our proposed depth-map-filling method, which is inspired by \cite{filling}. In Fig. \ref{fig:scanner_data} we see a typical depth map with undefined regions. We can observe that the whole scene is surrounded by a big undefined region. There are also small undefined areas near the warrior figure, caused by shadows or glossy surfaces of the scanned object. We divide undefined regions into two types (background and near-object) and treat them differently. We classify the undefined region as a background hole if some of its pixels are adjacent to the image border. Regions with no pixel adjacent to the image border are near-object holes.

\begin{figure}
\centering

\begin{subfigure}[b]{.30\linewidth}
\includegraphics[width=\linewidth]{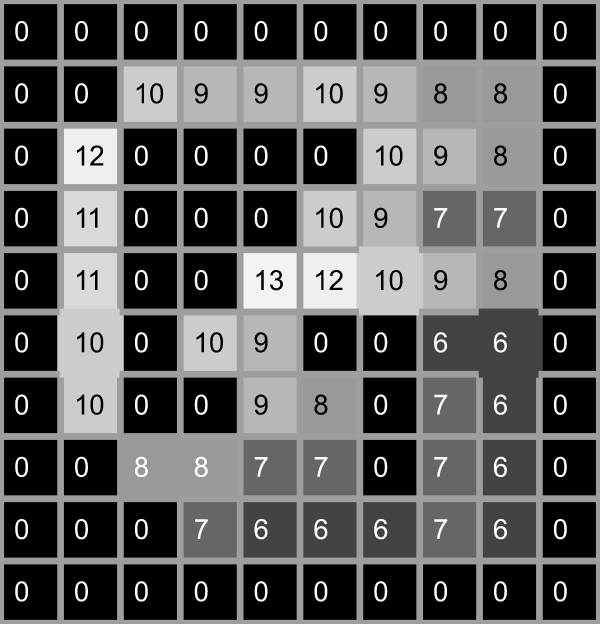}
\caption{}\label{fig:fill_input}
\end{subfigure}
\hfill
\begin{subfigure}[b]{.30\linewidth}
\includegraphics[width=\linewidth]{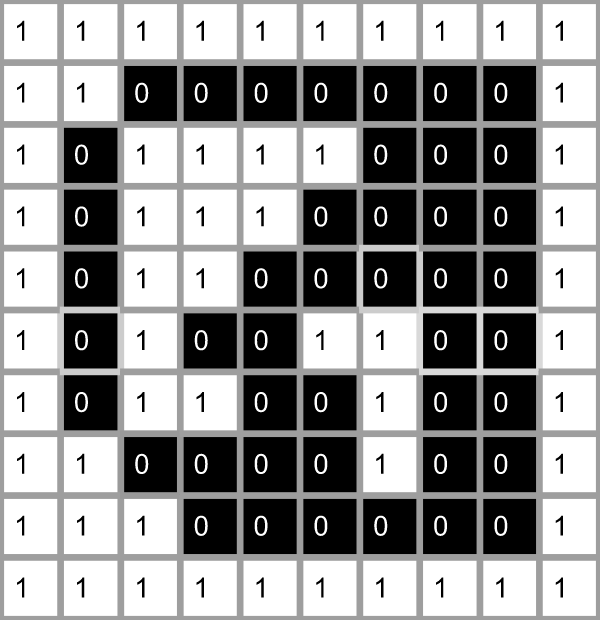}
\caption{}\label{fig:fill_holes}
\end{subfigure}
\hfill
\begin{subfigure}[b]{.30\linewidth}
\includegraphics[width=\linewidth]{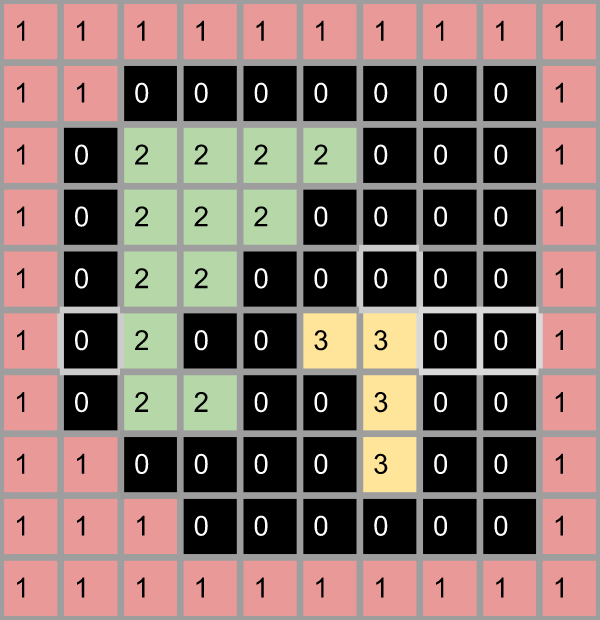}
\caption{}\label{fig:fill_labels}
\end{subfigure}
\hfill
\begin{subfigure}[b]{.30\linewidth}
\includegraphics[width=\linewidth]{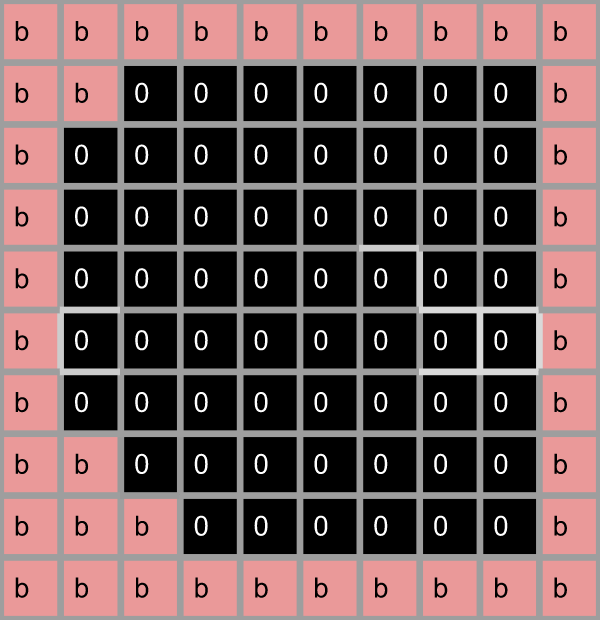}
\caption{}\label{fig:fill_back}
\end{subfigure}
\hfill
\begin{subfigure}[b]{.30\linewidth}
\includegraphics[width=\linewidth]{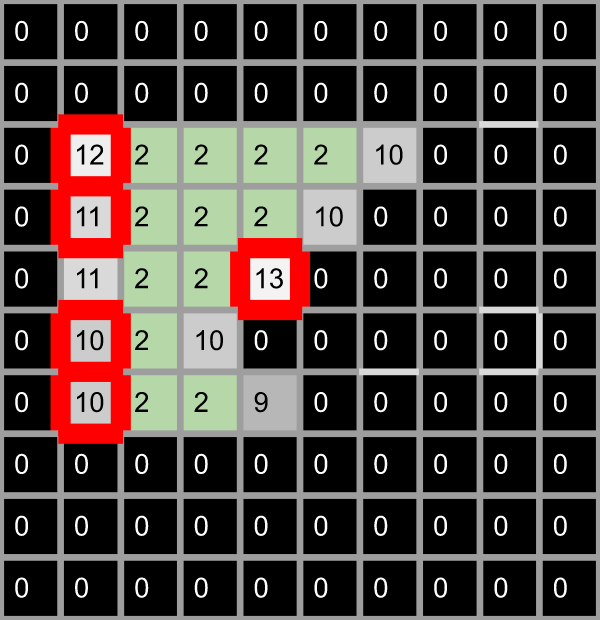}
\caption{}\label{fig:fill_max1}
\end{subfigure}
\hfill
\begin{subfigure}[b]{.30\linewidth}
\includegraphics[width=\linewidth]{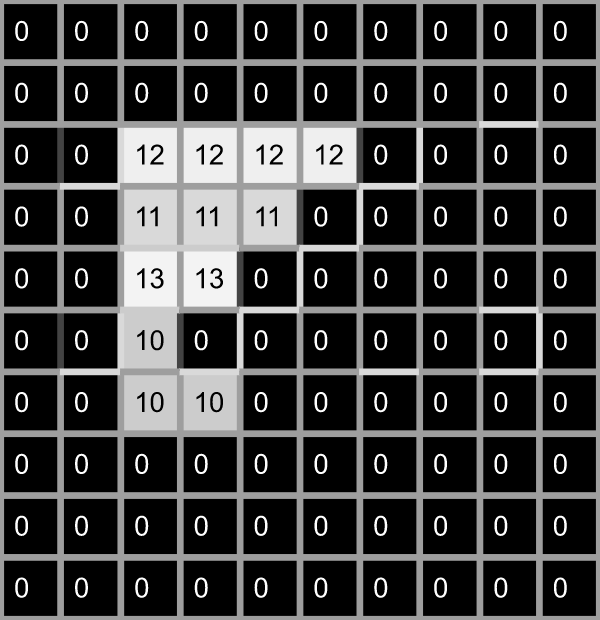}
\caption{}\label{fig:fill_rows1}
\end{subfigure}
\hfill
\begin{subfigure}[b]{.30\linewidth}
\includegraphics[width=\linewidth]{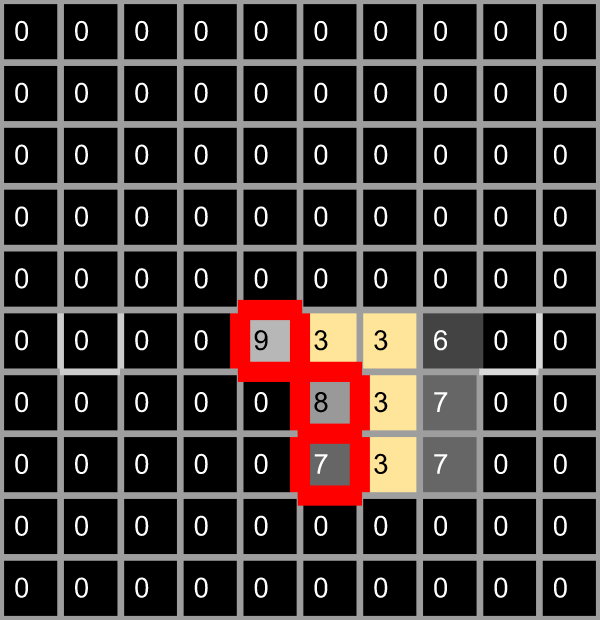}
\caption{}\label{fig:fill_max2}
\end{subfigure}
\hfill
\begin{subfigure}[b]{.30\linewidth}
\includegraphics[width=\linewidth]{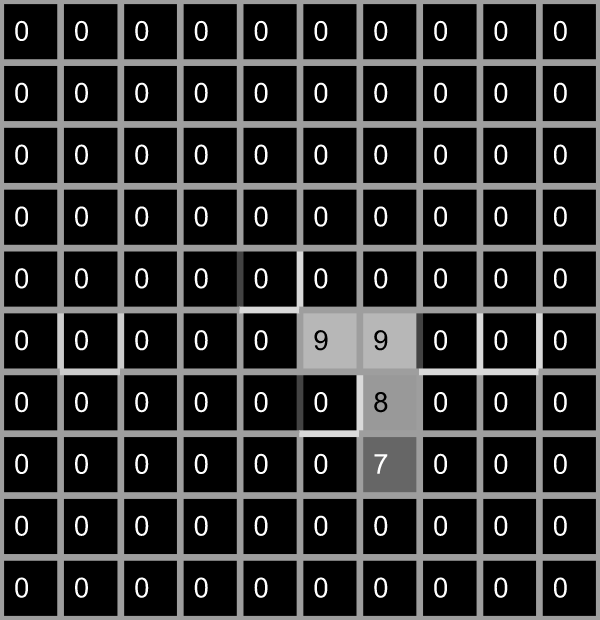}
\caption{}\label{fig:fill_rows2}
\end{subfigure}
\hfill
\begin{subfigure}[b]{.30\linewidth}
\includegraphics[width=\linewidth]{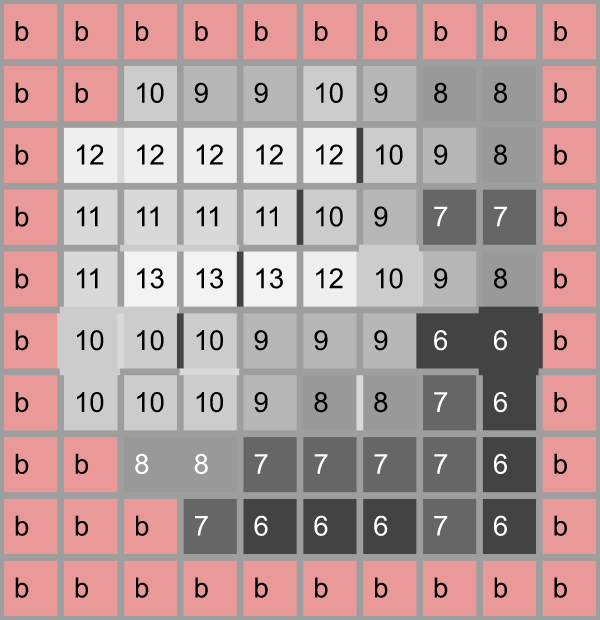}
\caption{}\label{fig:fill_result}
\end{subfigure}

\caption[Depth map filling]{
From the input (a), we generate the hole map (b) and label distinct holes with unique numbers (c). Background holes are filled with a single value (d). For the remaining holes, we determine the defined outer border pixels and find the maximum for each row (e) (g). Finally, we fill these holes row-wise (f) (h) and arrive at a fully defined depth map (i). 
}
\label{fig:filling_method}
\end{figure}

First, we generate a binary map of the holes. Next, we label every continuous hole region with a unique number. For each hole, if at least one border pixel has its label value, we treat it as a background hole and fill it with value $b$, discussed below. The remaining near-object holes are filled row-wise, using the maximum depth value from a defined border pixel in the given row, see Fig. \ref{fig:filling_method} for demonstration.

\begin{figure}
    \includegraphics[width=0.6\columnwidth]{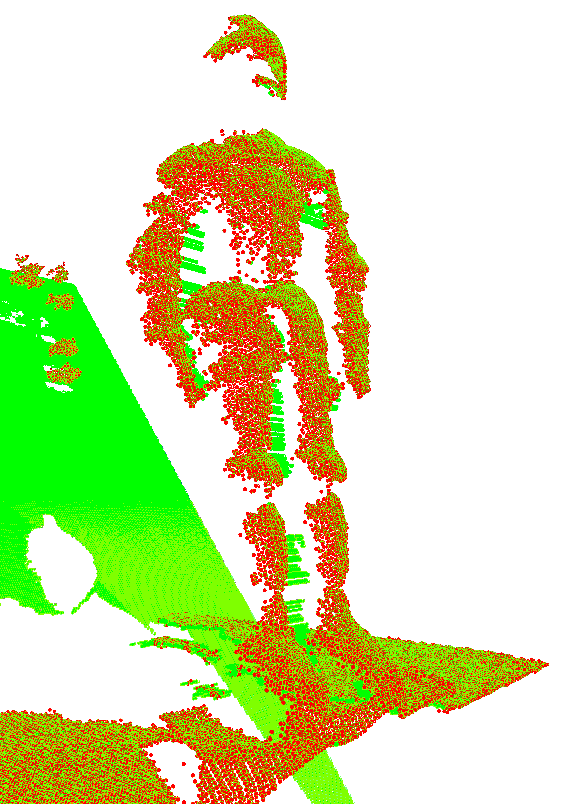}
    \centering
    \caption[Filled depth map detail]{Point cloud from filled depth map, defined points are in red, green points are filled by our proposed procedure. }
    \label{fig:filling_warior}
\end{figure}

Lastly, we need to consult the value $b$ used for background holes. We used the maximum depth value of all samples in the dataset, effectively forming a plane behind the scene, which gives some reference to the model during training. The filling method described above is not designed to preserve fine details or the most realistic object surface. We aim to keep the depth values of undefined areas stable and close to reality. The values do not need to be precise because, in the end, we set them back to an undefined state. Our experiments show that approximated hole-filling with synthetic background keeps the training process stable and the impact on the defined values is positive. Fig. \ref{fig:filling_warior} illustrates the filling method result.



\subsection{Texture Augmentation}
CNN models that we will implement use high-resolution intensity textures for extracting details of the scanned scene, such as edges or surface structures. 
There has to be a correlation between the intensity texture and depth map that we are up-sampling. For example, if some edge is not captured in the intensity texture, it would not be preserved precisely in the up-sampled depth map. On the contrary, if there is an edge in the intensity texture and the edge is missing in the depth map, it will also break the correlation of the images. 

We decided to modify the input intensity texture to preserve the correlation. First, we multiply the texture with the definition map from the depth map, to create matching undefined areas and then apply he same filling method described above. According to our experiments, this modification of intensity textures improved the training process. The leading cause of this phenomenon is probably the fact that we do not push the model to estimate useless depth values in the undefined regions.

\subsection{Depth Map Up-sampling}


As described in previous chapter, both FDSR and DKN models work with pre-upsampled depth maps. Input depth maps are already in a high-resolution from a simple expanding procedure and models only modify and enhance these inaccurate depth values. Models originally use bi-cubic interpolation, as they were designed for data from ToF devices \cite{kinect}. For images capture by structured-light scanners, nearest neighbor pre-upsampling is better at preserving sharp edges \cite{deconv} and we have confirmed this choice in our experiments.

\section{Training}

Training the model with pure $L_1$ loss showed weak results due to the model overfitting to a dominant plane surface in the image. Adding an edge based loss component as in \cite{fdsr} did not yield improvements. We have therefore opted to design a custom loss function which we call \textit{Object loss}.


\subsection{Loss Function}

The main mechanism behind our \textit{Object Loss} function is to give significantly higher attention to the depth map pixels belonging to the scanned object. The pixels of flat ground surrounding the object in the scene also contribute to the loss but with smaller weight. First, we compute an object map which is a binary matrix that for every pixel contains the value 1 if the pixel belongs to the object and 0 if not. Using an object map, we compute weighted $L_1$ error that gives weight $1$ to the object pixels and weight $0.01$ to the non-object pixels. The constant $0.01$ was determined empirically to achieve a stable training process. 


Obtaining the object map is derived from the composition of our scenes, which always comprise one object on the flat ground surface. If we find the geometric parameters of this ground surface, we can remove it and everything below it. We solved this issue by finding three pixels belonging to the flat ground which determine the searched plane. We assume that the biggest part of the scene contains the ground. That means the ground pixel depth values should be closer to the depth map mean value than the object depth values. Based on this assumption, we compute for every depth map pixel its squared deviation from the image mean. Then we sort obtained distances and take the maximum from the lowest $5\%$ as a threshold for determining whether the pixel belongs to the plane.

\begin{figure}
\centering

\begin{subfigure}[b]{.49\columnwidth}
\includegraphics[width=\linewidth]{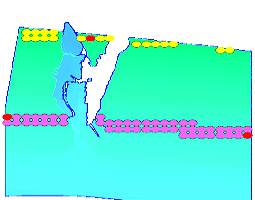}
\caption{}\label{fig:triangle_vertices}
\end{subfigure}
\hfill
\begin{subfigure}[b]{.49\columnwidth}
\includegraphics[width=\linewidth]{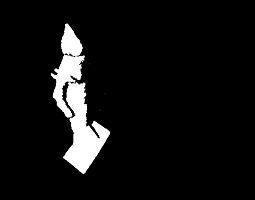}
\caption{}\label{fig:triangle_object}
\end{subfigure}

\caption[Object loss function]{(a) Yellow points represent the maximal-depth group of pixels of the grid, pink points represent the near-mean group of pixels of the grid, and the red points represent the chosen points to be vertices of the triangle determining scene ground plane. Figure (b) shows the output object map.}
\label{fig:triangle}
\end{figure}

When we visualized the group of $5\%$ mean-nearest pixels, see Fig. \ref{fig:triangle}, we observed that they were all placed near the horizontal line of the scene ground. The line was, in most samples, placed near the center of the scene ground. We choose the most left and right pixels along the y-axis as the 2 of the 3 searched pixels, forming a triangle base. To find a good plane, we must carefully choose the triangle's third vertex. The best would be a pixel near the scene's ground top or bottom border, far from the horizontal line from which we choose the triangle base vertices. Based on this reflection, we introduce a new group formed by the most distant pixels of the depth map. We empirically found that pixels with the depth value belonging to the highest $2\%$ of all depth map values belong to the scene ground near-top-border area. The third remaining pixel is chosen from this group by sorting along the y-coordinate and taking the middle pixel. This approach maximizes the length of the triangle sides. To speed up the whole process, we perform this search on down-sampled depth map with $20 \times 20$ resolution.


Finally, we generate the output object map by removing every depth value that fits the plane. The object map has been pre-computed while creating the dataset. Every sample then contains a filled LR depth map, HR intensity texture, HR definition map, HR object map, and the HR depth map as ground truth. The object loss function then uses the object map for computing the weighted error. 

\subsection{Architecture Adjustments}

One problem that we faced was our training hardware limit. DKN model, by its design, needs a large amount of GPU memory to work. Our depth maps have higher resolution than data from Kinect devices \cite{kinect}. 
To fit the model into GPU VRAM, we halved the number of filters in all convolution layers.
After a few experiments, we observed that the training process was stable with the reduced model. 
We rewrote the FDSR model in the PyTorch framework and validated both model implementations on the part of the NYUv2 dataset \cite{nyuv2}. 

We separate the $70\%$ of our 1200 samples as the training set with cross-validation and $30\%$ as the testing set. We train modified FDSR and DKN models, which both use Adam optimizer. The FDSR has a learning rate parameter set to $0.0005$, and the DKN model uses a $0.0001$. The FDSR model was trained in 1000 epochs. The DKN model was trained for 100 epochs. 
We use our custom object loss function for both models instead of the original $L_1$ loss.




\subsection{Data Post-processing}
DKN and FDSR models trained on our dataset are imperfect, especially in areas near the scanned object edges that often contain wrongly estimated outlier values. To preserve the image quality, we remove the incorrect values. After applying our model to the input data, we compute the point cloud and run the statistical outlier removal procedure from Open3D Library \cite{open3d}. This post-processing causes the loss of some points, but we value precision over quantity. The amount of the lost points vary but usually stays below $1\%$ of the input point cloud size. 

\section{Evaluation and results}

In this section we review possible applications of our trained models and examine their accuracy. 
Some metrics can be applied directly to the output depth map of our models, and other metrics can be applied to the 3D point cloud computed from the depth map. 

\subsection{Depth map metrics}
Authors of FDSR and DKN models use root mean squared error (RMSE) to evaluate their output depth map compared to the ground truth high-resolution one. This metric is unsuitable for us because our models are modified to give low attention to the biggest part of the whole image. Using RMSE on our data does not provide beneficial information about the output depth map quality because we are not interested in the error of the background plane but in the scanned object only. The more significant metric can be RMSE in combination with the object map, computing error only from a set of object pixels. We call this metric a \textit{Object RMSE}. We computed mean error values for the whole test dataset to quantitatively analyze our results using these depth map metrics. The test dataset contains 327 samples. 
Mean values were computed for FDSR, DKN, and simple nearest neighbor upscale as a baseline. All mentioned values stated in millimeters can be found in Tab. \ref{tab:rmse}.

\begin{table}
\centering
    \begin{tabular}{p{1cm}|p{1cm}|p{1.75cm}|p{1.5cm}}
        \hline
        \textbf{Method} & \textbf{RMSE} & \textbf{Object RMSE} & \textbf{Object Loss} \\
        \hline
        FDSR & 1.9537 & 4.4297 & 1.0427 \\
        DKN & 2.2696 & 6.9191 & 2.3058 \\
        Nearest & 3.0778 & 10.8156 & 7.0692 \\
        \hline
    \end{tabular}
\caption[Depth map metrics]{Mean depth map metrics values for the test split.}
\label{tab:rmse}
\end{table}

Depth map metrics described above do not provide direct information about resulting quality, because the cleaning process of outliers is done on an unstructured point cloud, so the RMSE metrics can not be applied afterward. However, these metrics can be helpful as a quick verification of the up-sampling model correctness. 
If we want to analyze the output more accurately, we need a metric that is applied after the outliers cleaning. 

\subsection{Point cloud metrics}

\begin{figure}
\centering

\begin{subfigure}[b]{.24\columnwidth}
\includegraphics[width=\linewidth]{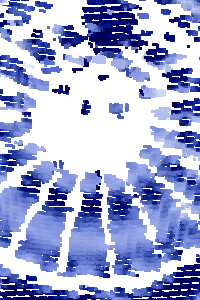}
\caption{}\label{fig:sample_lr}
\end{subfigure}
\hfill
\begin{subfigure}[b]{.24\columnwidth}
\includegraphics[width=\linewidth]{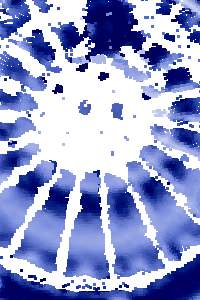}
\caption{}\label{fig:sample_hr}
\end{subfigure}
\begin{subfigure}[b]{.24\columnwidth}
\includegraphics[width=\linewidth]{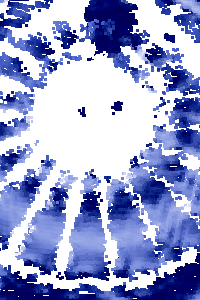}
\caption{}\label{fig:sample_dkn}
\end{subfigure}
\begin{subfigure}[b]{.24\columnwidth}
\includegraphics[width=\linewidth]{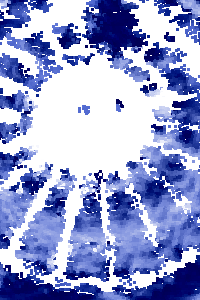}
\caption{}\label{fig:sample_fdsr}
\end{subfigure}

\caption[Sample point cloud analysis]{
Point cloud computed from (a) LR depth map, (b) ground truth HR depth map, (c) DKN and (d) FDSR model.
}
\label{fig:rotor_in}
\end{figure}

We first qualitatively analyze the resulting point clouds, see Fig. \ref{fig:rotor_in}. 
It is easier to recognize the differences in the shaded point cloud than in the raw depth maps. The baseline nearest neighbor method implies the grid-like structure of the scene surface. It can be seen that models, to some extent, eliminated this effect and estimated more accurate pixel values. 


\begin{figure*}[ht!]
\centering

\begin{subfigure}[b]{.1995\textwidth}
\includegraphics[width=\linewidth]{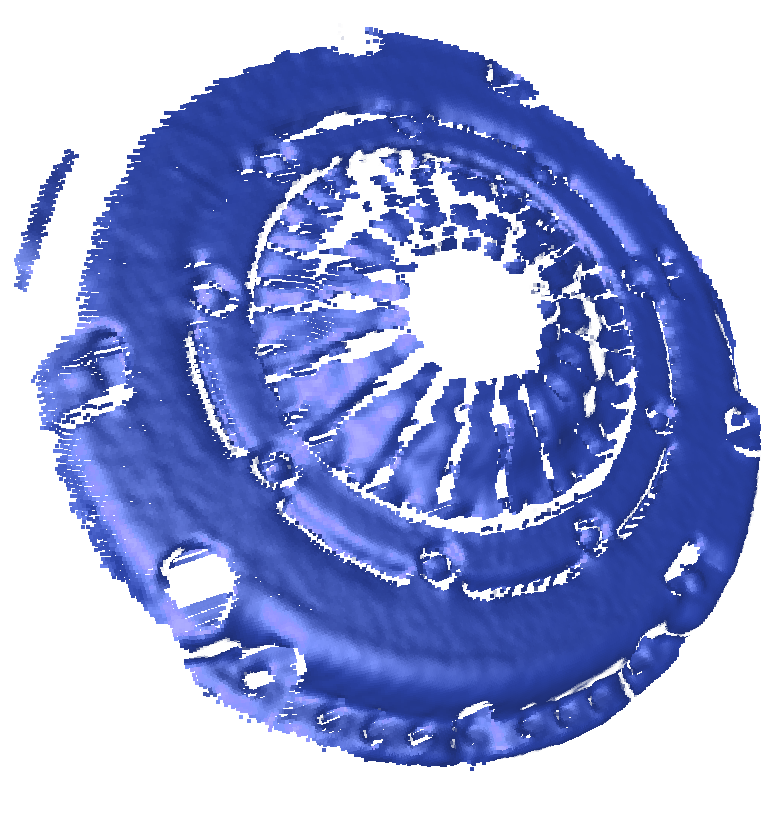}
\caption{}\label{fig:hausdorff_hr}
\end{subfigure}
\hfill
\begin{subfigure}[b]{.1995\textwidth}
\includegraphics[width=\linewidth]{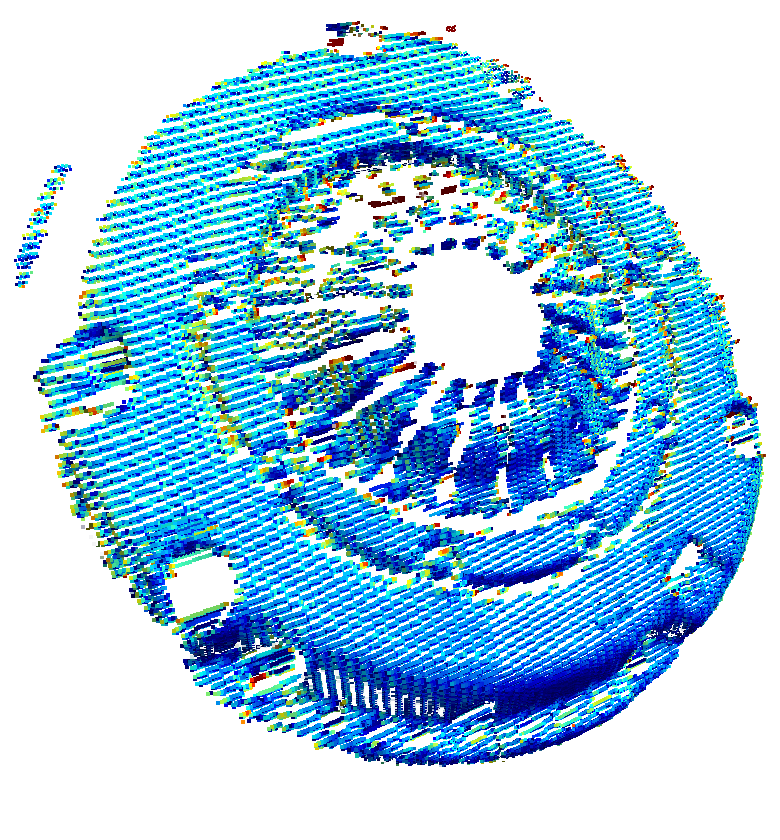}
\caption{}\label{fig:hausdorff_lr}
\end{subfigure}
\hfill
\begin{subfigure}[b]{.1995\textwidth}
\includegraphics[width=\linewidth]{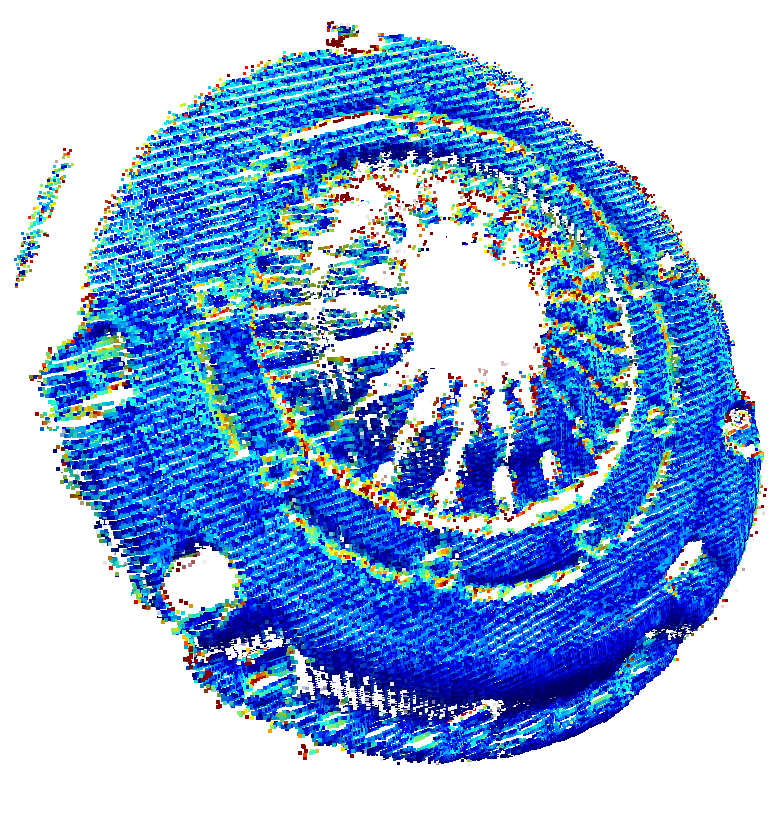}
\caption{}\label{fig:hausdorff_dkn}
\end{subfigure}
\hfill
\begin{subfigure}[b]{.1995\textwidth}
\includegraphics[width=\linewidth]{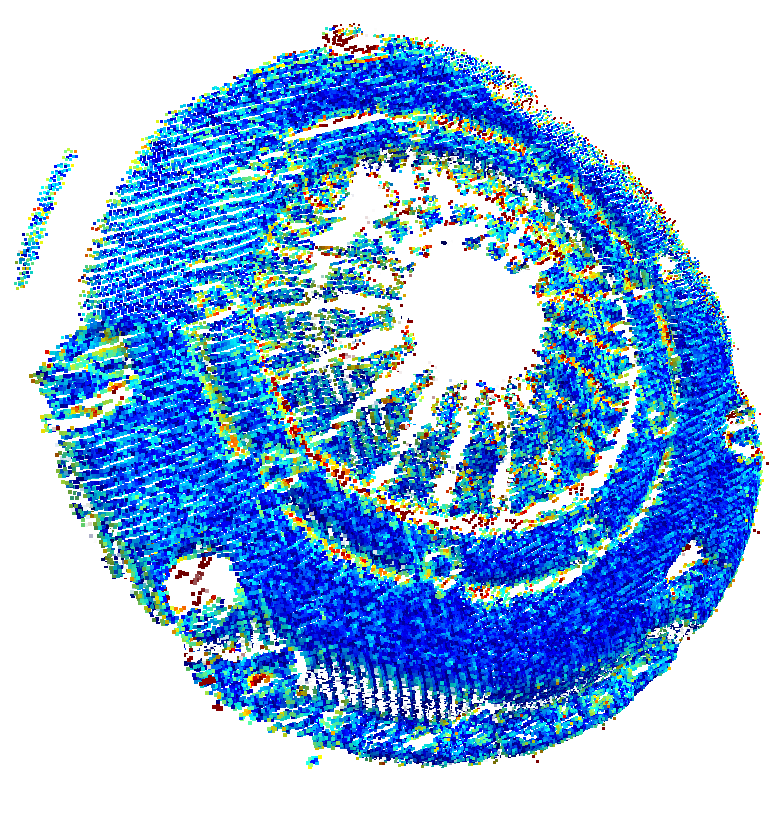}
\caption{}\label{fig:hausdorff_fdsr}
\end{subfigure}

\caption[Hausdorff distance analysis]{
Figure (a) shows the GT point cloud from HR depth maps. Haussdorf distances of points are mapped to color 
for point clouds calculated from (b) LR depth maps and outputs of (c) DKN model and (d) FDSR model. Red points are inaccurate, above the 2mm deviation. Distances in the accurate interval of 0 - 2mm are linearly mapped from blue through green to red.}
\label{fig:hausdorff}
\end{figure*}

To compare the resulting point clouds against the ground truth, we use the standard Haussdorf distance metric \cite{hausdorff}, see Tab. \ref{tab:hausdorff}. Mean values fit into 2 millimeters deviation, which we consider acceptable. Every up-sampling method contains outlier points with a large Hausdorff distance, with the nearest neighbor baseline achieving the highest maximum value by a large margin. On the other hand, it has the lowest mean values, suggesting that we should not rely solely on the statistical moments in our evaluation. Furthermore, we color the points based on the Hausdorff distance mapping, ranging from blue to red in Fig. \ref{fig:hausdorff}. We have used a threshold of 2 millimeters for the maximum error, meaning all red points can be considered inaccurate. Based on these visualizations, we can conclude that FDSR and DKN models provide a more precise object surface but have significant errors along the object edges. The nearest neighbor output also contains inaccurate points near edges and the whole object surface is significantly less accurate. The FDSR and DKN clearly outperform the simple nearest neighbor up-sampling method, which proves the usability of these models.

\begin{table}
\centering
    \begin{tabular}{p{1cm}|p{1cm}|p{1cm}|p{1cm} }
        \hline
        \textbf{Method} & \textbf{min} & \textbf{max} & \textbf{mean} 
        \\
        \hline
        FDSR &  0.0000 & 9.8178 & 0.4650 
        \\
        DKN &  0.0000 & 10.3815 & 0.4449 
        \\
        Nearest &  0.0000 & 29.5586 & 0.4144 
        \\
        \hline
    \end{tabular}
\caption[Point cloud metrics]{Statistical moments of Hausdorff distance applied on output point clouds from up-sampling methods against the ground truth point clouds from the HR depth maps.}
\label{tab:hausdorff}
\end{table}

\subsection{Time measurements}

One of the motivations for the depth map super-resolution task was speeding up possible down-stream tasks, such as outlier removal. This can be achieved by filtering in low resolution and up-sampling back using our models. To evaluate the efficiency of our solution, we measured the times original filtering pipeline for different resolutions, see Tab. \ref{tab:pipeline} for results. Processing time tends to grow exponentially while increasing the depth map resolution.

\begin{table}
\centering
    \begin{tabular}{p{1.75cm}|p{1cm}|p{1cm}|p{1.2cm}|p{1.2cm}}
        \hline
        Resolution & 140x200 & 560x800 & 1120x800 & 1680x1200 \\
        \hline
        Time [s] & 0.054 & 0.068 & 0.091 & 0.184 \\
        \hline
    \end{tabular}
\caption[Pipeline processing time]{Measured time of the depth map filtering. The pipeline accepts three different resolutions, therefore the time for the lowest resolution was estimated by fitting a second-degree polynomial function to known high-resolution values.}
\label{tab:pipeline}
\end{table}

To analyze the possible processing time improvement by depth map super-sampling, we measured the time of FDSR \cite{fdsr} and DKN \cite{dkn} models. On depth maps with $140 \times 200$ resolution, the average running time across $10$ samples was $0.007$ seconds for FDSR and $0.634$ seconds for DKN. As the FDSR model is about 100 times faster than the DKN model, it can be used for speeding up the pipeline. The advantage of using DKN is the higher precision of results. Therefore it is suitable for the depth map quality improvement, which was our second motivation. It allows to up-sample observations from cheaper devices with lower resolution.

\section{Conclusion}

In conclusion, two modified CNN models have been trained, FDSR and DKN, for super-sampling depth maps. Our enhancements in data pre-processing and custom loss function showed promising improvements. The models outperformed the nearest neighbor method in terms of accuracy and surface smoothness. The FDSR model offers faster processing time, while the DKN model provides higher precision results. The models can be used to improve the quality of low-resolution depth maps and potentially speed up the pipeline with further optimization. 
The proposed approach offers benefits such as reducing the processing time of a pipeline by 
performing various processing steps at the lower resolution and upsampling the resulting depth map or increasing the resolution of a point cloud captured in lower resolution by a more affordable device.


\section*{Acknowledgment}

This publication is the result of support under the Operational Program Integrated Infrastructure for the project: Advancing University Capacity and Competence in Research, Development and Innovation (ACCORD, ITMS2014+:313021X329), co-financed by the European Regional Development Fund. The work presented in this paper was carried out in the framework of the TERAIS project, a Horizon-Widera-2021 program of the European Union under the Grant agreement number 101079338.

\bibliographystyle{plain}
\bibliography{literature} 

\end{document}